\documentclass[conference]{IEEEtran}
\IEEEoverridecommandlockouts
\usepackage[numbers,sort&compress]{natbib}
\usepackage{amsmath,amssymb,amsfonts}
\usepackage{algorithmic}
\usepackage{graphicx}
\usepackage{textcomp}
\usepackage{xcolor}
\usepackage{booktabs}
\usepackage{multirow}
\usepackage{stfloats}
\usepackage{blindtext}
\def\BibTeX{{\rm B\kern-.05em{\sc i\kern-.025em b}\kern-.08em
    T\kern-.1667em\lower.7ex\hbox{E}\kern-.125emX}}

\newcommand{\abdu}[1]{\textbf{\textcolor{red}{[Abdu: #1]}}}
\newcommand{\joey}[1]{\textbf{\textcolor{green}{#1}}}
\newcommand{\abdulrahman}[1]{\textbf{\textcolor{blue}{#1}}}

\begin{document}

\title{From SQL Errors to Concept Gaps: An AI-Powered Knowledge Graph Analytics Platform for Diagnostic Feedback\\
{\footnotesize \textsuperscript{}}
\thanks{\copyright~2026 IEEE. Personal use of this material is permitted. Permission from IEEE must be obtained for all other uses, in any current or future media, including reprinting/republishing this material for advertising or promotional purposes, creating new collective works, for resale or redistribution to servers or lists, or reuse of any copyrighted component of this work in other works.}
}

\author{
\IEEEauthorblockN{Abdulrahman AlRabah, Weijian Zhou, Xing Gao, Abdussalam Alawini}
\IEEEauthorblockA{Siebel School of Computing and Data Science, University of Illinois Urbana-Champaign\\
\{alrabah2, zhou86, xinggao6, alawini\}@illinois.edu}
}


\maketitle

\begin{abstract}
This innovative practice full paper describes an AI-powered knowledge graph platform that connects SQL errors to conceptual gaps in undergraduate and graduate database systems courses. Students learning Structured Query Language (SQL) frequently struggle with semantic errors that reflect conceptual misunderstandings rather than syntax mistakes. A query may execute yet return incorrect results due to gaps spanning related concepts; misusing NATURAL JOIN in place of an explicit subquery reflects intertwined misunderstandings of JOIN, GROUP BY, and HAVING. Autograding systems detect correctness but provide surface-level feedback without connecting errors to the conceptual structure of the course. Educational knowledge graph research has shown the value of structured concept representations for curriculum analysis and adaptive learning, but these approaches have not been applied to diagnosing SQL misconceptions from student submissions. We present a platform that automatically extracts course concepts and relations from instructional materials, links them to student submission traces through a graph database, and classifies errors at the concept level. We evaluate the platform across two database systems courses at two universities, one using real student submissions and one using simulated submissions, through an expert study with five participants and an automated evaluation using an LLM as a judge. Results show that 95.7\% of extracted nodes were rated as at least somewhat valid and 63.8\% of triplets were rated fully correct. Expert feedback confirmed that the generated graphs align with instructor mental models and that mapping errors to course concepts provides actionable diagnostic insight; evaluating impact on student learning remains future work.
\end{abstract}


\section{Introduction} 
SQL is one of the most widely taught topics in database systems courses  given its fundamental role as the standard language for querying and manipulating relational data. However, SQL is challenging for many students because it requires a different kind of reasoning than more procedural programming tasks; instead of describing each step of a process, SQL requires students to specify (declare) the result they want. Previous research has shown that this shift creates persistent conceptual difficulties, specifically around SQL concepts, including Join, Grouping and Aggregation, Filtering (WHERE), and Nested Queries \cite{ahadi2016students,taipalus2018errors,taipalus2019expect}. More recent works have also analyzed student SQL homework and problem-solving patterns at scale, showing that student difficulties tend to be systematically observable through repeated submission behavior rather than confined to isolated failed attempts. \cite{poulsen2020insights,yang2021analyzing,yang2023uncovering, zhao2022edukg}.

These difficulties are not captured well by existing autograding systems that mostly focus on query correctness. A student may submit a query that runs but still reflects flawed reasoning, or may repeatedly edit a query through trial and error without understanding the underlying concept they are missing. While autograders help instructors evaluate submissions at scale, simple binary feedback is often too coarse to explain why a solution failed or what misconception produced the failure. Prior work on SQL tutoring has therefore emphasized interactive tools, personalized guidance, and richer feedback generation beyond simple grading \cite{brusilovsky2010learning,kleiner2024enhancing}. However, most existing systems still focus on individual attempts rather than maintaining an explicit representation of the conceptual structure of the course.

That limitation matters in real classrooms because student errors are rarely isolated. A failed submission may reflect a gap in a single concept, but it may also arise from a misunderstanding of how several related concepts fit together. Meanwhile, the key course information of a course is often distributed across lecture slides, practice problems, referenced solutions, and submission logs. Knowledge graphs offer a promising way to organize course concepts, enhancing curriculum analysis, adaptive learning, and interpretable learning support, as highlighted by recent educational research \cite{abu2024systematic,aytekin2024ace,canal2024educational, zhao2022edukg}. However, prior work has also noted that constructing and maintaining educational knowledge graphs manually is labor-intensive, motivating the development of automated and AI-assisted approaches \cite{abu2024systematic,aytekin2024ace}.

Motivated by these gaps, we present an AI-powered knowledge graph analytics platform for SQL education. The system automatically extracts course concepts and relations from instructional materials, links them to assessment questions and submission traces, and uses this structure to contextualize student errors at the concept level. Its architecture separates execution-based grading from model-assisted semantic interpretation, while asynchronously synchronizing learner evidence into a graph database for instructor-facing analytics and student-facing progress feedback. 

This work makes the following key contributions.
\begin{itemize}
\item It introduces an automated pipeline for constructing a course knowledge graph from instructional materials and assessment content.
\item It combines execution-based grading with LLM-based semantic feedback to classify SQL errors and connect them to relevant course concepts.
\item It supports concept level analytics over student submissions, enabling both diagnostic feedback for students and instructor-facing views of recurring concept gaps and misconception patterns.
\end{itemize}

\section{Related Work} 
Prior works on SQL learning system have analyzed student errors through semantic taxonomies, misconception analysis, and large-scale analyses of student submissions. \citet{ahadi2016students} show that students make recurring semantic mistakes across different SQL query types rather than only isolated syntax slips.  \citet{taipalus2018errors} further categorize errors and complications in SQL query formulation and argue that these difficulties are systematic and instructionally meaningful. \citet{miedema2022identifying} deepen this line of work by examining SQL misconceptions and their causes, highlighting that many student errors reflect incomplete or incorrect mental models rather than simple carelessness. \citet{poulsen2020insights} show that student difficulties are visible in how solutions evolve across attempts, emphasizing the value of analyzing submission traces rather than only final correctness outcomes. Together, these studies motivate our focus on semantic SQL errors, but beyond prior analyses that primarily study errors as submission-level phenomena, our work connects them to an explicit course concept graph and student submission traces for concept-level interpretation.

SQL tutoring and autograding systems have increasingly moved beyond correctness checking toward richer formative feedback. \citet{brusilovsky2010learning} integrate interactive tools and personalization for SQL learning, showing the value of guided practice environments. \citet{kleiner2024enhancing} generate structural hints and partial credit by comparing student and reference parse trees, demonstrating that autograded feedback can reflect query structure rather than only final output. \citet{yousef2025begrading} present BeGrading, a fine-tuned LLM-based system for automated grading and feedback on programming assignments, demonstrating that LLM generated evaluations can approach human grading consistency while reducing educator workload. Within SQL specifically, \citet{manikani2025sql} develop an LLM powered SQL autograder using fine-tuned open-source models that achieves 96.77\% grading accuracy and provides structured feedback on incorrect submissions, demonstrating the scalability of LLM-based evaluation in university settings. \citet{lai2025enhancing} present a GPT 4o based feedback system for SQL assignments that uses structured prompt engineering and predefined rubrics to evaluate both syntax and semantics, achieving approximately 91\% feedback accuracy across 80 students. \citet{alrabah2025codelens} propose CodeLens, a generative AI framework that detects semantic SQL errors and produces non-revealing feedback intended to guide students without giving away the solution. Their framework combines a general purpose LLM for semantic error detection with a fine-tuned model trained on human curated feedback to filter out solution revealing content. Evaluation across two semesters showed that students receiving AI supported feedback required fewer submission attempts, particularly on complex queries involving subqueries and conditional logic. However, these systems all operate at the individual submission-level without maintaining a persistent representation of how errors relate to broader course concepts and dependencies. Our system improves by reusing submission-level feedback as evidence inside a knowledge-graph analytics pipeline rather than treating each attempt as a standalone tutoring event.

Educational knowledge graph research has shown that structured representations of concepts and relations can support curriculum analysis, recommendation, and adaptive learning. \citet{abu2024systematic} emphasizes that educational knowledge graphs are useful for concept mapping, personalized learning, and instructional organization, while also noting that graph construction and maintenance remain difficult bottlenecks. \citet{aytekin2024ace} propose an expert-in-the-loop approach for building educational knowledge graphs with prerequisite relations, showing that AI assistance can reduce manual effort while preserving interpretability. Similarly, \citet{alatrash2025inferring} combine document based, graph based, and text based features through a voting algorithm to infer prerequisite relationships without labeled data. \citet{alrabah2026instructor} propose InstructKG, a framework that extracts concepts and infers prerequisite and compositional dependencies from lecture materials. Our platform adopts a similar dependency structure and extends it by connecting the concept graph with student submission traces and error classification for concept-level diagnosis that can in turn support personalized learning.

A key gap in the literature is that prior work on SQL errors, feedback, and educational knowledge graphs has largely progressed separately. Existing SQL tutoring work is strong at identifying or explaining query errors, while prior educational knowledge graph work is strong at organizing concepts and prerequisites. Our work bridges these strands together by linking student submission traces, semantic SQL error types, extracted course concepts, and concept relations in one unified graph-based representation. This unified representation allows the system to trace students errors back to specific course concepts and their prerequisites, giving both instructors and students a structured view of where conceptual gaps and misconceptions originate.

\section{Methodology} 






\begin{figure*}[t]
\centering
\includegraphics[width=0.95\textwidth]{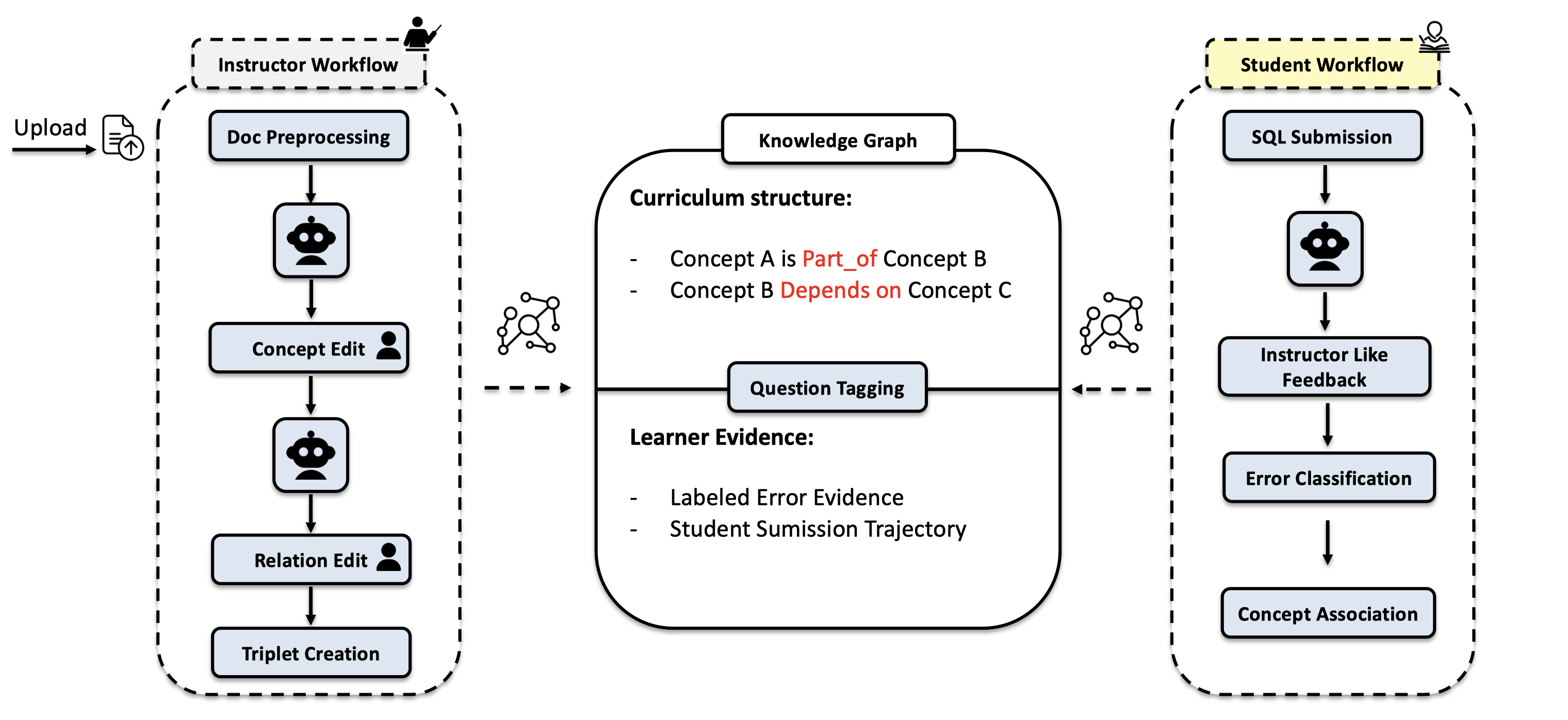}
\caption{End-to-end architecture of the submission evaluation and graph analytics platform. The online path performs low-latency auto-grading and controlled semantic feedback generation, while the asynchronous path synchronizes stored attempts and extracted concepts into Neo4j for concept-level analytics, reconciliation, and dashboard support.}
\label{fig:system_architecture}
\end{figure*}

\subsection{Dataset and Data Preparation}
Our platform operates on two synchronized data spaces: a course-content space for assessment definitions and instructional materials, and a student-attempt space for submission traces. The course-content space is built from structured JSON question files exported from the university's online assessment platform. For autograded SQL items, the question JSON is extended with the prompt, executable schema, test data, expected output, and a reference query, which together define a deterministic grading target. The student-attempt space is stored in PostgreSQL through the \texttt{submissions}, \texttt{submission\_errors}, and \texttt{vector\_embeddings} tables defined in the backend ORM.

This separation is necessary because the system must support both low-latency grading and cross-attempt analytics. We therefore normalize all records by \texttt{student\_id}, \texttt{assessment\_id}, and \texttt{question\_id}, and preserve timestamps, status labels, feedback, and embeddings for each stored attempt. During ingestion, question text is cleaned to remove interface-only artifacts such as trailing \texttt{query.sql} instructions, while student histories are appended rather than overwritten. The resulting schema provides a stable join key across grading, error aggregation, and knowledge-graph synchronization.

\subsection{System Architecture}\label{sec:architecture}
Figure~\ref{fig:system_architecture} illustrates the end-to-end platform architecture, organized around two workflows that feed into a shared knowledge graph. The instructor workflow (left) handles course content ingestion, where uploaded lecture materials are preprocessed, and LLM assisted extraction proposes candidate concepts and relations. An instructor can review and edit the extracted concepts and relations before they are finalized as triplets (concept-relation-concept edges, e.g., \texttt{INNER JOIN ---> Part Of ---> JOIN}) and synchronized into the knowledge graph. The student workflow (right) processes SQL submissions through execution-based grading and LLM assisted semantic feedback, followed by error classification and concept association, which links each identified error to relevant nodes in the knowledge graph. The central knowledge graph unifies both sides, storing curriculum structure (prerequisite and compositional relationships between concepts) alongside learner evidence (labeled error records and student submission trajectories). A question tagging layer bridges the two workflows by associating each assessment question with its target concepts in the graph, which enables the system to trace student errors back to specific course concepts and their dependencies.

\subsection{Assessment Pipeline and Auto-Grading}
The online evaluation module addresses a core instructional requirement: returning reliable feedback immediately after each attempt. The frontend sends \texttt{submission}, \texttt{question\_id}, \texttt{student\_id} , and \texttt{assessment\_id} to \texttt{/api/evaluate}, after which the backend loads the corresponding question JSON and detects the submission language. For SQL submissions, the grader constructs a temporary database instance, executes the provided schema and test tuples, runs the student query, and compares the produced result with the expected output. This execution-first design yields three system-wide status labels: \texttt{correct}, \texttt{compilation\_error}, and \texttt{semantic\_error}.

Semantic feedback is generated only when execution succeeds but the result still differs from the reference behavior. In that case, the system first uses a general-purpose LLM-based semantic validator to analyze the cleaned student query against the problem statement and hidden reference solution. The output is then passed to a fine-tuned model trained on a labeled dataset of approximately 123 human-curated non-revealing feedback examples designed to mimic instructor-style feedback that does not disclose the solution, which filters the explanation to remove any content that could disclose the correct answer before it returns to the learner. This two-stage design matches the implementation goal of preserving useful semantic guidance while reducing the chance of leaking the answer directly. Each attempt is stored in PostgreSQL together with its status, feedback, and vector embedding, so later analytics operate on the same records used during grading.

The technical advantage of this module is that it combines deterministic correctness checking with controlled model-based explanation. Deterministic execution avoids ambiguous grading decisions, while the structured feedback layer converts non-correct submissions into reusable pedagogical signals such as \texttt{JOIN}, \texttt{GROUP BY}, \texttt{WHERE}, alias, and aggregate-function errors. Those typed labels are later reused by student and instructor insight endpoints, which means the online feedback module also serves as the data source for higher-level learning analytics.

\subsection{Knowledge Graph Construction}

\begin{figure}[t]
\centerline{\includegraphics[width=0.47\textwidth]{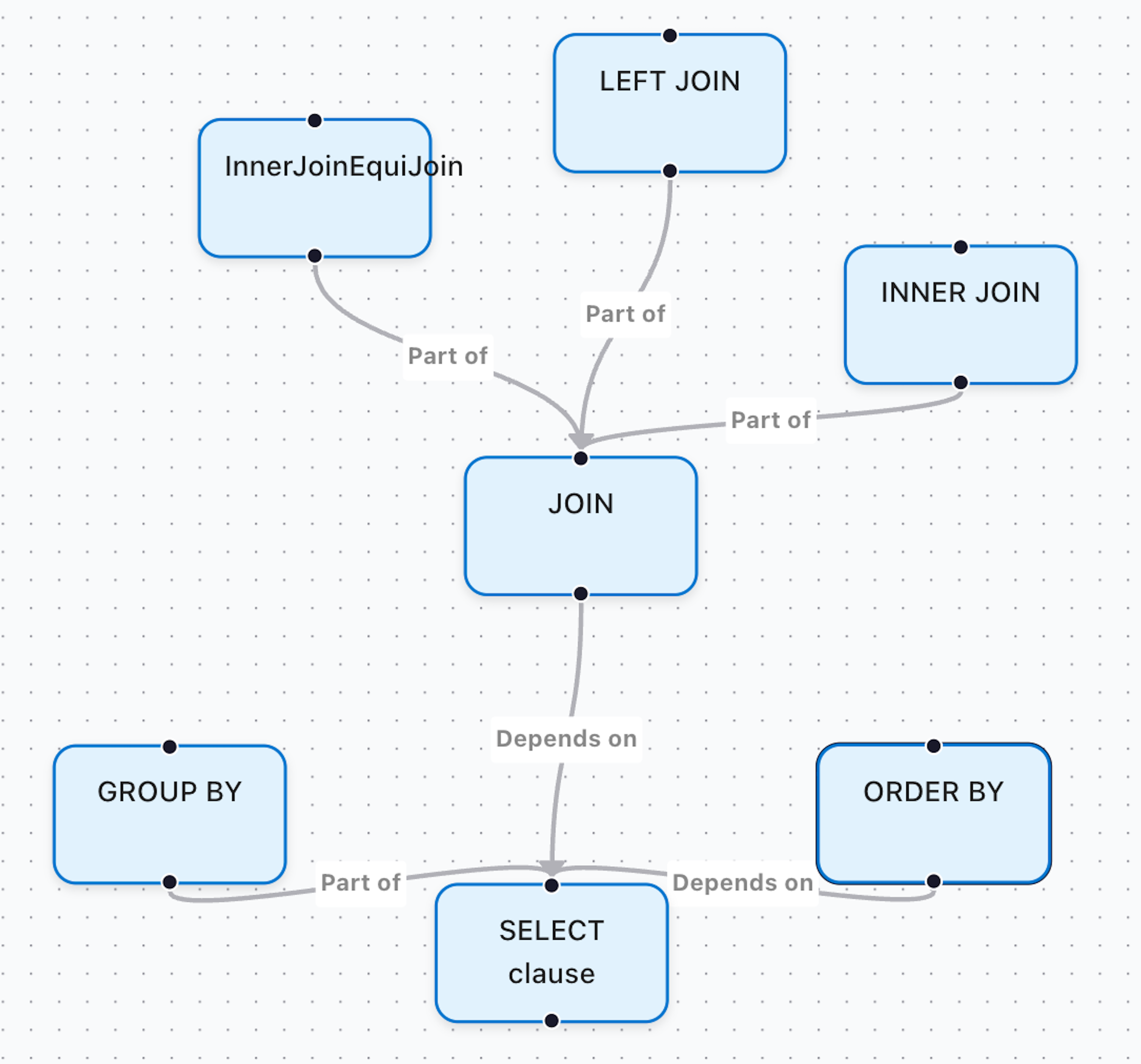}}
\caption{Subgraph of the course knowledge graph showing concept relationships. \texttt{PART\_OF} edges indicate compositional relationships and \texttt{DEPENDS\_ON} edges indicate prerequisite dependencies}
\label{fig:concept_graph}
\end{figure}

The knowledge-graph module is designed to unify curriculum structure and student submission activity in a single queryable representation. Figure~\ref{fig:concept_graph} presents the conceptual schema used to connect assessment structure, question-level concept targets, student attempts, and typed error evidence. For readability, the schema omits several implementation-specific intermediary nodes used during graph synchronization while preserving the core methodological semantics. On the curriculum side, uploaded materials are converted into text, chunked, filtered, and processed by an LLM-based extraction pipeline to propose concept nodes and relations. Because the same concept may appear under slightly different surface forms across lecture slides and assessments, the extracted concepts are canonicalized through the \texttt{canonical\_concepts} table before synchronization into Neo4j. This canonicalization step is essential for preventing fragmented nodes that would weaken downstream aggregation.

On the learner side, graph updates are executed asynchronously after the attempt has been stored. A background task reads \texttt{student\_id}, \texttt{assessment\_id}, \texttt{question\_id}, and associated error labels from PostgreSQL, then writes them into Neo4j as a typed path connecting student, assessment, question, and error entities. This staged write path is motivated by two constraints: the graph should reflect the latest classroom activity, but graph I/O should not slow down the response returned by the evaluation endpoint. The asynchronous design satisfies both constraints while preserving a direct link between raw submission logs and graph nodes. To ensure quality, concept extraction and question tagging are validated through the instructor review workflow described in Section~\ref{sec:architecture}, while the error-to-concept associations derived from the feedback module are based on the same typed labels that instructors can inspect and correct through the platform's reconciliation interface.

The resulting graph supports concept-aware reasoning that flat tables cannot express easily. Student-question nodes can be connected to curriculum concepts, concept-concept relations, and error categories in the same structure, enabling instructors to inspect whether a recurring error is localized to one assessment item or propagates across related concepts. Because the graph is built from normalized database records rather than separately curated summaries, the resulting analytics remain traceable to observed student behavior.

Figure~\ref{fig:concept_graph} shows a subgraph of the resulting concept structure for a database systems course. Concepts are connected through two relation types: \texttt{PART\_OF} edges capture compositional relationships (e.g., INNER JOIN, LEFT JOIN, and InnerJoinEquiJoin are all part of JOIN), while \texttt{DEPENDS\_ON} edges capture prerequisite relationships (e.g., JOIN and ORDER BY both depend on SELECT clause, and GROUP BY is part of SELECT clause). These relations are proposed by an LLM-based on the uploaded lecture materials and finalized through instructor review, as described in Section~\ref{sec:architecture}.

To connect student submissions to this concept structure, the platform performs question tagging. When a student submits a query and receives an error classification from the feedback module, the system identifies the assessment question being attempted and the error type produced. Given these inputs, the platform determines candidate concepts from the knowledge graph that are associated with the question and builds edges linking the question node to the relevant concept nodes. This allows each student error to be traced through the graph to the specific concepts involved and, through the dependency edges, to the prerequisite concepts that may require reinforcement.

\subsection{Triplets, Reconciliation, and Progress Analytics}
Ontology refinement and progress estimation are implemented as the final analytics layer. Stored attempts, question-concept links, and proposed concept relations are combined to support dashboard-facing insights. To move beyond simple concept co-occurrence, the platform includes a reconciliation workflow that proposes candidate concept triplets and normalizes accepted relations into graph-safe edge types such as \texttt{DEPENDS\_ON} and \texttt{PART\_OF}. The motivation for this step is that concept extraction alone identifies relevant vocabulary but does not capture the instructional dependencies between concepts needed for interpretable ontology growth and instructor review. By inserting a reviewable reconciliation stage, the system allows instructors to accept, reject, or modify proposed relations before they enter the graph, ensuring the ontology can grow without treating every automatically proposed relation as ground truth.


Progress analytics are computed from concept-linked question records and correctness histories, while the reconciliation workflow is maintained as a separate ontology-refinement layer. For each concept \(c\), class-level completion is defined as
\begin{equation}
P_c = \frac{N^{\text{correct}}_c}{N^{\text{possible}}_c}\times 100\%,
\end{equation}
where \(N^{\text{correct}}_c\) is the number of correct concept-linked attempts and \(N^{\text{possible}}_c\) is the number of eligible student-question opportunities associated with that concept. Student-level mastery is computed as the fraction of tracked concepts for which the learner has demonstrated at least one correct outcome, and low-mastery concepts are surfaced as personalized focus targets.

This module provides the main technical advantage of the overall system: it converts isolated grading events into longitudinal evidence about concept mastery. The resulting summaries are exposed through progress-insight endpoints and consumed by teacher and student dashboards. Instructors receive cohort-level views of persistent concept gaps and common error types, while students receive individualized mastery estimates anchored in their own submission history. These metrics are grounded in stored attempts and graph-linked concept evidence, ensuring they remain auditable rather than purely heuristic.

\begin{table}[t]
\small
\caption{Methodology Components and Their Roles}
\begin{center}
\begin{tabular}{|p{2.4cm}|p{4.9cm}|}
\hline
\textbf{Component} & \textbf{Role in Pipeline} \\
\hline
Question Bank & Supplies prompts, schemas, test data, expected outputs, and reference SQL. \\
\hline
Execution Grader & Executes SQL in temporary MySQL and assigns \texttt{correct}, \texttt{compilation\_error}, or \texttt{semantic\_error}. \\
\hline
Semantic Feedback & Uses \texttt{gpt-4o} plus fine-tuned no-reveal filtering for concise error explanations. \\
\hline
KG Construction & Extracts course concepts and relations from lecture materials using \texttt{gpt-5-mini}. \\
\hline
Persistence Layer & Stores attempts, statuses, errors, and embeddings in PostgreSQL. \\
\hline
Graph Sync & Ingests student attempt traces into Neo4j as typed student-assessment-question-error paths. \\
\hline
Progress Analytics & Computes concept completion, mastery percentage, and focus concepts for dashboards. \\
\hline
\end{tabular}
\label{tab:method_components}
\end{center}
\end{table}


\section{Evaluation}

\subsection{Study Design} 

This study was conducted under an approved IRB protocol. We recruited five participants, one instructor and three teaching assistants from one course and one instructor from the other, to evaluate the platform's outputs across two database systems courses at two universities. Both courses enroll a mix of undergraduate and graduate students with varying levels of prior SQL experience. Each participant reviewed a knowledge graph generated from their own course materials, including up to 27 sampled concept nodes and up to 23 sampled triplets, along with five student submission cases showing the system's predicted concepts and misconceptions. For Course A, the platform processed 7 lectures and 4 assessment questions with real student submissions from 5 consented students. For Course B, the platform processed 5 lectures and 5 homework questions with 5 simulated student submissions, as real submission data was not available for that course.
The evaluation uses two metrics. Metric 1 assesses KG correctness and interpretability through item level ratings of node validity and triplet accuracy on a three-point scale, followed by six overall Likert questions and open ended feedback. Metric 2 assesses error classification accuracy through per submission ratings of concept relevance, misconception validity, helpfulness, and ranking accuracy on five-point scales, followed by eight overall questions and open-ended feedback. To test whether automated evaluation can approximate expert judgment, we administered the same quantitative rubric to a large language model given identical inputs and measured agreement using Spearman and Pearson correlation.

\subsection{Evaluation Results}

\begin{table*}[t]
\centering
\setlength{\tabcolsep}{16pt}
\caption{Expert and LLM evaluation of knowledge graph correctness and error classification across two courses. Item-level values are reported as percentages (\%). Overall quality and error classification are reported as pooled means.}
\label{tab:eval-combined}
\begin{tabular}{l cc cc cc}
\toprule
\multirow{2}{*}{\textbf{Metric}}
& \multicolumn{2}{c}{\textbf{Course A ($n{=}4$)}}
& \multicolumn{2}{c}{\textbf{Course B ($n{=}1$)}}
& \multicolumn{2}{c}{\textbf{Combined ($n{=}5$)}} \\
\cmidrule(lr){2-3} \cmidrule(lr){4-5} \cmidrule(lr){6-7}
& \textit{Node} & \textit{Triplet}
& \textit{Node} & \textit{Triplet}
& \textit{Node} & \textit{Triplet} \\
\midrule
\multicolumn{7}{l}{\textit{Metric 1: KG Correctness -- Expert Evaluation (\%)}} \\
\addlinespace[2pt]
Rated valid (2)    & 54.5 & 62.0 & 88.5 & 69.6 & 60.9 & 63.8 \\
Rated somewhat (1) & 40.2 & 31.0 & 11.5 & 21.7 & 34.8 & 28.7 \\
Rated invalid (0)  &  5.4 &  7.0 &  0.0 &  8.7 &  4.3 &  7.4 \\
Mean (0 to 2)      & 1.49 & 1.55 & 1.88 & 1.61 & 1.57 & 1.56 \\
\addlinespace[4pt]
\multicolumn{7}{l}{\textit{Metric 1: KG Correctness -- LLM Evaluation (\%)}} \\
\addlinespace[2pt]
Rated valid (2)    & 81.5 & 71.4 & 100.0 & 63.6 & 90.7 & 67.4 \\
Rated somewhat (1) & 18.5 & 28.6 &   0.0 & 36.4 &  9.3 & 32.6 \\
Rated invalid (0)  &  0.0 &  0.0 &   0.0 &  0.0 &  0.0 &  0.0 \\
Mean (0 to 2)      & 1.81 & 1.71 &  2.00 & 1.64 & 1.91 & 1.67 \\
\midrule
\multicolumn{7}{l}{\textit{Metric 1: KG Overall Quality (1 to 5 scale, pooled mean)}} \\
\addlinespace[2pt]
Expert              & \multicolumn{2}{c}{3.25} & \multicolumn{2}{c}{4.75} & \multicolumn{2}{c}{3.75} \\
LLM                 & \multicolumn{2}{c}{4.00} & \multicolumn{2}{c}{4.00} & \multicolumn{2}{c}{4.00} \\
\midrule
\multicolumn{7}{l}{\textit{Metric 2: Error Diagnosis and Concept Association (1 to 5 scale, pooled mean)}} \\
\addlinespace[2pt]
Expert              & \multicolumn{2}{c}{3.06} & \multicolumn{2}{c}{4.13} & \multicolumn{2}{c}{3.28} \\
LLM                 & \multicolumn{2}{c}{3.96} & \multicolumn{2}{c}{4.36} & \multicolumn{2}{c}{4.16} \\
\bottomrule
\end{tabular}
\end{table*}

Table~\ref{tab:eval-combined} summarizes the evaluation results across both courses. Expert evaluation of the knowledge graph showed that 95.7\% of nodes were rated as at least somewhat valid, with 60.9\% rated as both valid and meaningful. Triplet accuracy followed a similar pattern, with 63.8\% rated fully correct and only 7.4\% rated invalid. Overall KG quality averaged 3.75 out of 5, while error diagnosis averaged 3.28 out of 5. Across the four Course A evaluators, agreement on the basic validity of items was high, with 88.5\% pairwise agreement on whether a given node or triplet was valid (a nonzero rating). Course B consistently received higher ratings across all metrics, though it reflects a single evaluator.

\noindent\textbf{Automatic Evaluation. }To complement the expert study, we conducted a separate automated evaluation using GPT-4o-mini as an LLM judge. We provided the model with the same concept nodes, triplets, and error classification items that experts reviewed, along with the identical scoring rubrics and scale definitions. The model scored each item independently with no access to expert responses. The LLM rated 90.7\% of nodes as fully valid compared to 60.9\% by experts and never assigned a score of 0 to any item, indicating a consistent positive bias. Triplet accuracy was more closely aligned at 67.4\% versus 63.8\%. For overall KG quality, the LLM produced a uniform 4.00 across both courses, while expert ratings ranged from 3.25 to 4.75, suggesting the LLM does not capture variance in perceived quality across different course contexts. On error classification, the LLM rated both courses higher than experts, with the gap being larger for Course A (3.96 vs 3.06) than Course B (4.36 vs 4.13). These findings suggest that LLM evaluation produces directionally similar judgments but with a positive bias, and may serve as a scalability tool for initial screening rather than a replacement for expert review.

\subsection{Qualitative Findings}

For Metric 1, participants across both courses found the generated knowledge graph to be broadly representative of their course structure. One instructor noted that ``the graph closely captured the mental model that I have about the structure of the SQL part of the course,'' while another stated it ``matches my mental model to a large extent.'' Participants suggested that tooltips with concept summaries, references to corresponding lecture slides, and the ability to edit nodes and edges would further improve interpretability. The most common structural issue was reversed edge directions, which evaluators from both courses flagged for specific triplets.

Looking at Metric 2, participants found the system effective at identifying the general conceptual area a problem engages, with one evaluator noting it helps "instructors or TAs quickly orient themselves to what the question is testing." The system's handling of aggregation errors was a noted highlight, where one participant observed that the GROUP BY label "directly points an instructor to ask the student: what does each group represent," making the issue immediately actionable. The Course B instructor described the labels as "very useful" overall. Participants noted that label granularity could be improved in some cases, with the appropriate level of detail being context-dependent across assignments and exams.

\section{Discussion}

\begin{figure}[t]
\centerline{\includegraphics[width=0.47\textwidth]{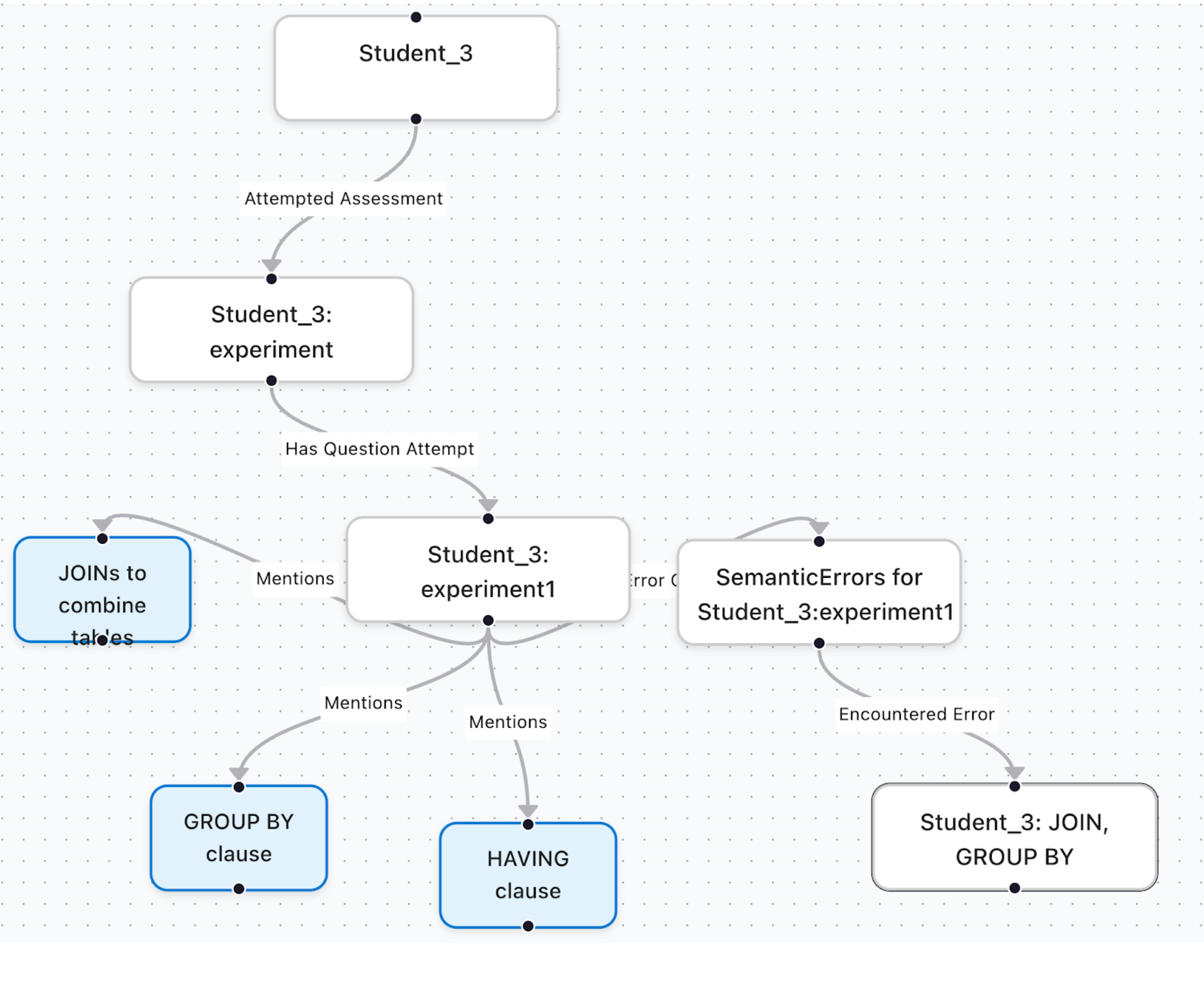}}
\caption{Knowledge graph subgraph for Course\_A Student\_3 showing the connection from student submission to assessment, course concepts (blue), and classified semantic errors.}
\label{fig:student_graph}
\end{figure}

The evaluation results reveal several patterns worth examining. Course B consistently received higher ratings across all metrics. However, this difference should be interpreted with caution, as Course B reflects the judgment of a single evaluator, whereas Course A aggregates ratings from four evaluators with potentially varying expectations and levels of familiarity with the generated graph. Inter-rater variability is a known factor in expert evaluation studies, and the higher consistency in Course B's scores may partly reflect the absence of such variability rather than a meaningful difference in system performance.

A recurring theme in the qualitative findings was the issue of edge direction and edge type. Participants from both courses flagged cases where prerequisite arrows were reversed, and one instructor questioned whether the distinction between DEPENDS\_ON and PART\_OF is necessary at all. Automated extraction from unstructured lecture materials can identify that two concepts co-occur, but determining which one is the prerequisite requires pedagogical reasoning that current LLMs do not always handle reliably. A practical mitigation is to incorporate an instructor review step for proposed edges before they enter the graph, which the platform's reconciliation workflow already supports.

Expert feedback also highlighted that while the system correctly identifies the general concept area involved in a student error, the labels often lack the granularity needed to pinpoint the specific mistake. For example, labeling an error as "JOIN" does not distinguish between misuse of NATURAL JOIN, missing join conditions, or confusion between inner and outer joins. This tradeoff is inherent in automated concept tagging: broader labels generalize across students and problems but sacrifice diagnostic precision. Future work could explore hierarchical labeling, where a coarse label like "JOIN" is paired with a finer sub-label derived from the specific query difference.

The LLM judge evaluation showed that automated scoring produces directionally similar results to expert judgment but with a consistent positive bias. The LLM never assigned a score of 0 and tended to rate items higher than experts. This pattern is consistent with known tendencies of LLMs to be lenient evaluators \cite{zheng2023judging}. For practical deployment, this means LLM-based evaluation may be useful for identifying clearly problematic items that score low even under a lenient judge, but it should not replace expert review for distinguishing between adequate and strong quality.
A key design motivation of the platform is enabling concept-level error diagnosis that can unlock personalized learning pathways through the connection between student errors and the course concept graph. Figure~\ref{fig:student_graph} illustrates this for a concrete case. Student\_3 submitted an incorrect query using NATURAL JOIN instead of the required subquery with explicit joins. The system classified the error as involving JOIN and GROUP BY, and the question node links to three course concepts: JOINs to combine tables, GROUP BY clause, and HAVING clause. Because these concept nodes are connected through DEPENDS\_ON and PART\_OF edges in the curriculum graph, the system can surface not only what the student got wrong but also which prerequisite concepts they may need to revisit. This kind of structured feedback moves beyond telling a student their query is incorrect and toward explaining where in the conceptual map their understanding breaks down.

\section{Limitations}

While the results are encouraging, the system has only been tested with SQL based assessments. Transferring this approach to other programming languages or non-technical domains would require adapting the extraction pipeline and the error classification schema. However, the underlying architecture, which separates concept extraction, error classification, and graph based analytics, is domain agnostic in design and could be extended with appropriate prompt engineering and domain specific ontologies.
\section{Conclusion and Future Work}

We presented an AI-powered knowledge graph analytics platform that connects SQL error diagnosis with structured course concept representations to support personalized learning. The platform automatically extracts concepts and their dependencies from instructional materials, classifies student errors at the concept level through a combination of execution-based grading and LLM assisted semantic feedback, and links both into a shared graph database that supports instructor and student facing analytics. Evaluation across two database systems courses at two universities showed that 95.7\% of extracted concept nodes were rated as at least somewhat valid by expert evaluators, and that mapping errors to course concepts was perceived as actionable for identifying student knowledge gaps. Qualitative feedback confirmed that the generated graphs align with instructor mental models and that concept level error labels help instructors quickly identify what a question is testing and where students struggle.

Several directions remain for future work. First, the granularity of error labels could be improved through hierarchical concept tagging, where coarse labels such as JOIN are paired with finer sub-labels that distinguish specific misconceptions. Second, the platform could be extended to support domains beyond SQL by adapting the concept extraction pipeline and error classification schema to other programming languages or technical subjects. Third, a longitudinal deployment with real student submissions would enable evaluation of the platform's impact on learning outcomes over a full semester rather than through expert review alone. Finally, integrating adaptive recommendation into the student facing dashboard, where the system suggests specific review materials based on identified concept gaps, would strengthen the connection between diagnosis and intervention.

\bibliographystyle{IEEEtranN}
\bibliography{references}

\end{document}